\documentclass{article}
\usepackage{spconf,amsmath,graphicx,hyperref}

\usepackage{enumitem}
\setlist{nosep, leftmargin=14pt}

\usepackage{mwe} 
\usepackage[ruled, linesnumbered]{algorithm2e} 
\DontPrintSemicolon
\usepackage{amsfonts} 
\usepackage{caption}

\title{Multimodal Interactive Lung Lesion Segmentation: A Framework For Annotating PET/CT Images Based on Physiological and Anatomical Cues}
\name{
    \begin{tabular}{@{}c@{}}
Verena Jasmin Hallitschke$^{1*}$ \quad
Tobias Schlumberger$^{1*}$  \quad 
Philipp Kataliakos$^{1*}$ \quad 
Zdravko Marinov$^{1,2}$\\
Moon Kim$^3$ \qquad 
Lars Heiliger$^3$ \qquad
Constantin Seibold$^{1,2}$ \qquad 
Jens Kleesiek$^3$ \qquad
Rainer Stiefelhagen$^1$
\end{tabular}
}
\address{
$^1$ Institute for Anthropomatics \& Robotics (IAR), Karlsruhe Institute of Technology, Germany 
\\
$^2$ HIDSS4Health - Helmholtz Information and Data Science School for Health, Karlsruhe, Germany
\\
$^3$ Institute for AI in Medicine (IKIM), University Hospital Essen, Germany
}

\begin{document}
%
\maketitle
\begin{abstract}
    Recently, deep learning enabled the accurate segmentation of various diseases in medical imaging. These performances, however, typically demand large amounts of manual voxel annotations. This 
    tedious process for volumetric data 
    becomes 
    more complex when not all required information is available in a single imaging domain as is the case for 
    PET/CT data. 

    We propose a multimodal interactive segmentation framework that mitigates these issues by combining anatomical and physiological cues from PET/CT data. Our framework utilizes the geodesic distance transform to represent the user annotations and we implement a novel ellipsoid-based user simulation scheme during training. We further propose two annotation interfaces and conduct a user study to estimate their usability. We evaluated our model on the in-domain validation dataset and an unseen PET/CT dataset. 
    We make our code publicly available \href{https://github.com/verena-hallitschke/pet-ct-annotate}{here}. 

\end{abstract}
\makeatletter{\renewcommand*{\@makefnmark}{}
\footnotetext{* denotes equal contribution.}\makeatother}
\section{Introduction}





With the proliferation of large-scale annotated datasets, supervised deep learning (DL) models have achieved state-of-the-art performance in vision tasks such as classification, object detection, and semantic segmentation~\cite{russakovsky2015imagenet, patel2020upsurge, dosovitskiy2020image, ronneberger2015u}. However, the success of DL models is attributed to large manually annotated datasets. The annotation 
is especially demanding for volumetric medical data due to the difference among patients, variability in the shape and appearance of the pathology, and the three-dimensional nature of the data~\cite{rother2021assessing}.

Interactive segmentation models mitigate these issues by accelerating the annotation and iteratively improving the label quality as the user guides the segmentation model using foreground and background interactions, e.g., clicks or scribbles, to correct its predictions leading to high-quality masks.
The voxelwise annotations are reduced to a few user interactions, which alleviates the burden of manual annotation. For this reason, interactive models have gained popularity in 3D segmentation for medical image analysis~\cite{wang2018deepigeos, luo2021mideepseg, wang2020uncertainty, roth2021going, sommer2011ilastik, cho2021deepscribble, li2021wdtiseg}.

\setlength{\abovecaptionskip}{2pt}
 \setlength{\belowcaptionskip}{-15pt}
\begin{figure}[t]
    \centering
    \includegraphics[width=\columnwidth]{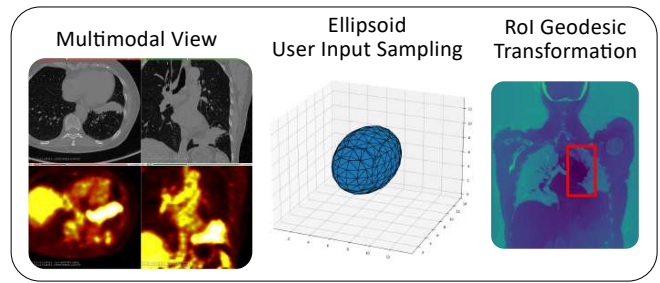}
    \caption{
    We introduce an 
    interactive segmentation framework containing paired multimodal views
    (left).
    We propose a novel ellipsoid-based user simulation scheme during training (middle) and we use only the RoI when computing the geodesic transform (right) to save computational time.}
    \label{fig:overview}
\end{figure}

\begin{figure*}[t]
    \centering
    \includegraphics[width=0.8\textwidth]{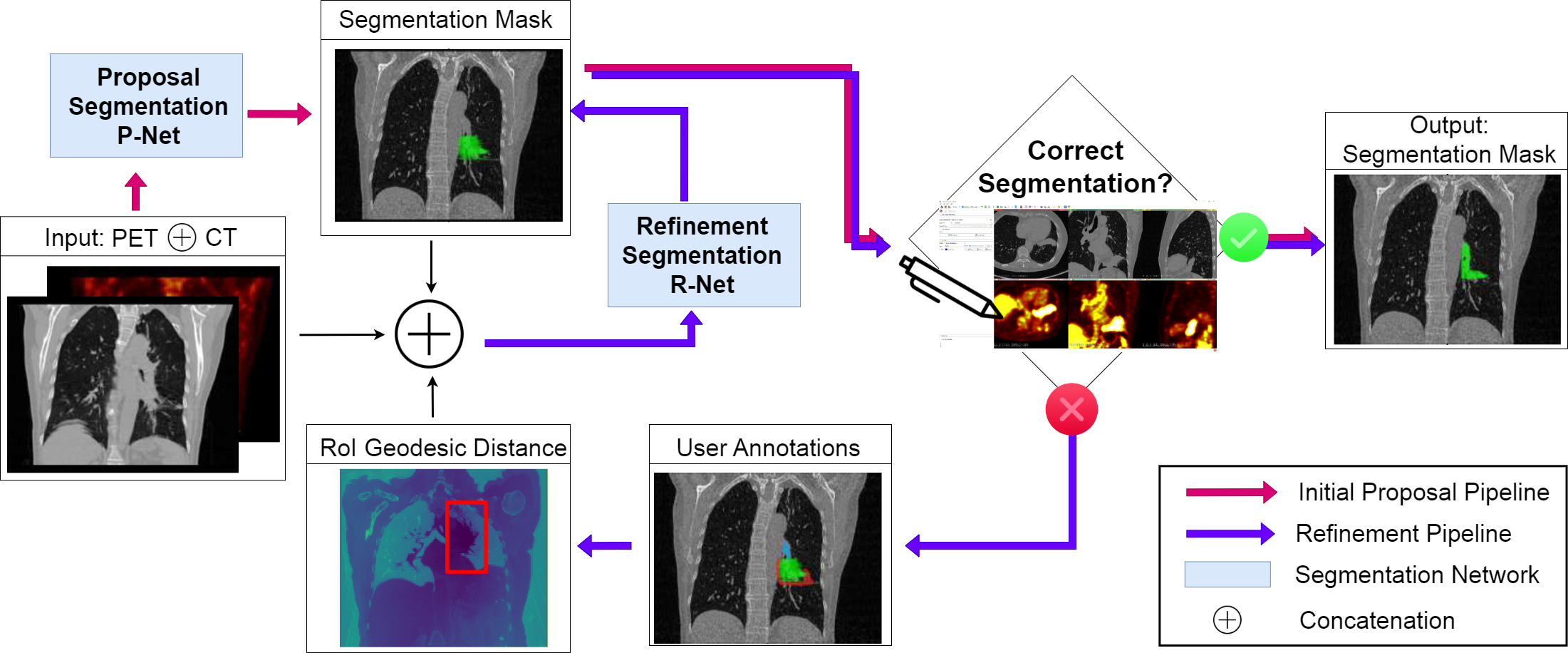}
    \caption{The workflow of our proposed method. Blue boxes indicate the segmentation models which are adapted from Wang et~al.~\cite{wang2018deepigeos} to multimodal inputs.}
    \label{fig:workflow}
\end{figure*}

Interactive segmentation models have shown remarkable performance in various imaging modalities, such as Computed Tomography (CT)~\cite{luo2021mideepseg, roth2021going}, Magnetic Resonance Imaging (MRI)~\cite{wang2018deepigeos, luo2021mideepseg, wang2020uncertainty}, and Ultrasound~\cite{li2021wdtiseg}. However, interactive models have not been yet applied in a multimodal setting. Including multiple imaging domains increases the diversity of the training data and provides more information about the examined pathology~\cite{AutoPET}. For example, in PET/CT imaging, the PET emphasizes regions with high
metabolic uptake, which is typical for a plethora of tumors~\cite{krause2013fdg}, whereas the CT is often obtained at a much higher resolution providing more detailed information about the affected anatomy~\cite{townsend2004pet}. CT scans also indicate scars from treated tumors, which do not appear notably in the PET images 
\cite{chen2004application}.

To this end, we extend the paradigm of interactive segmentation to multimodal imaging. More concretely, we show how to train a multimodal interactive segmentation model on the recently released PET/CT AutoPET dataset~\cite{AutoPET}. We examine two ways to present the two modalities to the annotators and conduct a user study to estimate how each view influences the performance. Additionally, we propose a novel scheme for simulating user interactions during training, which is based on random ellipsoids, and encode the interactions using an optimized RoI geodesic distance transform. We evaluate our method on the unseen Lung-PET-CT-Dx dataset~\cite{LungPETCT} to show its generalization performance.

\section{Method}
The approach proposed in this work consists of two parts: a user interface which is used to interact with the system and annotate PET/CT volumes, and a backend model, which is intended to speed up the annotation process of the users by proposing an initial rough segmentation mask, and refining this mask iteratively with the help of user inputs in the form of foreground and background scribbles as seen in Figure \ref{fig:overview}. 

We adapt the DeepIGeoS interactive segmentation model \cite{wang2018deepigeos} and extend it to two imaging modalities, e.g., PET/CT. DeepIGeoS \cite{wang2018deepigeos} consists of a CNN-based proposal model P-Net and a refinement model R-Net which share the same architecture. P-Net is first trained end-to-end for automatic segmentation. Then, R-Net is trained to refine P-Net's predictions by concatenating the original input, P-Net's prediction, and the user's foreground and background annotations as R-Net's input during training. 

Figure \ref{fig:workflow} shows the workflow of our framework which is composed of several steps. Firstly, before any user interactions, an initial proposal segmentation is produced by the model. This proposal segmentation is then presented to the annotator along with the PET and CT images in the interface.
The user decides whether to accept the current segmentation or to add corrective background and/or foreground scribbles in over- and undersegmented areas. The scribbles are encoded using a RoI geodesic distance transform and concatenated to the PET/CT volume as a joint input. The joint input is then fed to the model which produces a new refined segmentation and shows it to the user. This refinement cycle continues until the user is satisfied with the current quality of the segmentation and submits it to the system.

\subsection{User Interface}

\begin{figure*}[t]
    \centering
    \includegraphics[width=0.9\textwidth]{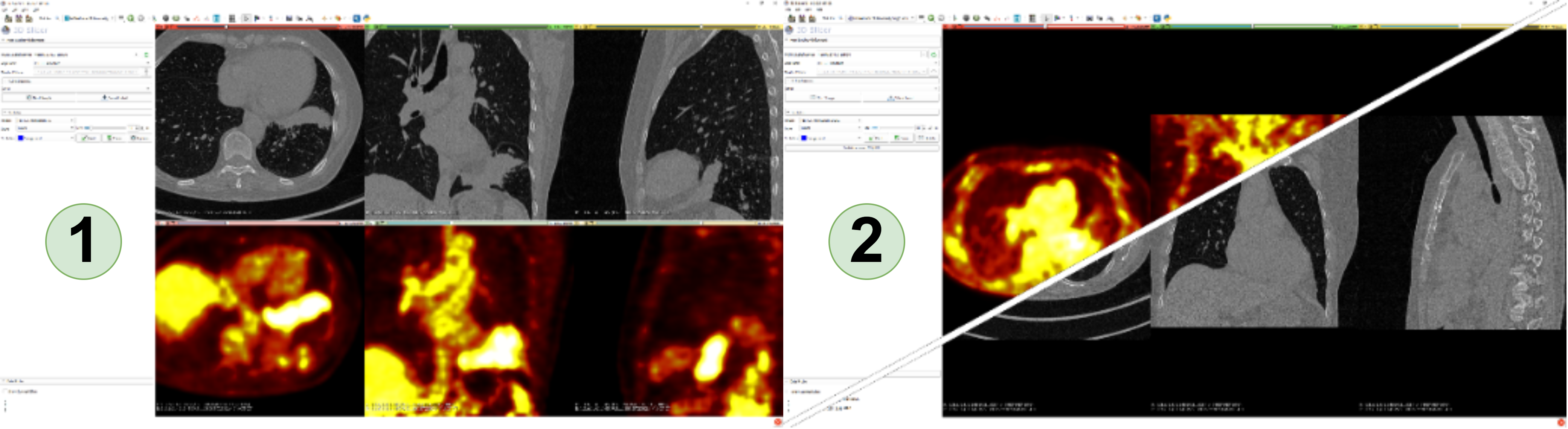}
    \caption{The interface of the annotation tool during an annotation process involving both PET and CT images.}
    \label{fig:interface}
\end{figure*}

The user interface is implemented as a plugin of 3D Slicer~\cite{Slicer}, utilizing MONAI~Label~\cite{MONAILabel} to integrate human interactions and model inference.
Figure \ref{fig:interface} shows the two options for the user interface of the tool during an annotation process. In the first option (1), the user is presented with both image modalities at the same time, seeing them side by side. The mouse is also duplicated on the same location of both the PET and CT modalities to ease the annotation and the  slice views are linked, i.e., both views zoom and pan when either of them changes. This allows the simultaneous processing of the information from both modalities, e.g., assessing the general location of a tumor via the PET image and determining more precise boundaries of tissue in the CT image. The second option for the interface (2) is to show only one modality at a time, where the user can switch the modality using a button. This way, the user can focus his attention entirely to one modality. We evaluate the usability and performance gain from each interface with a user study in Section \ref{subsec:evaluation}.

\subsection{Model Architecture and Training}

\textbf{Architecture.} We adopt the DeepIGeoS~\cite{wang2018deepigeos} model for our interactive segmentation framework and extend it to multimodal PET/CT imaging data. Our model consists of two networks: P-Net which is used for creating a proposal segmentation, and R-Net which iteratively refines the proposed segmentation with the help of user inputs until the user agrees with the quality of the prediction. The input for P-Net is the concatenated PET and CT volumes, while R-Net additionally appends the foreground /background annotations of the user and the previous prediction to the input as seen in Figure \ref{fig:workflow}.

\begin{figure}[b]
    \centering
    \includegraphics[width=0.98\linewidth]{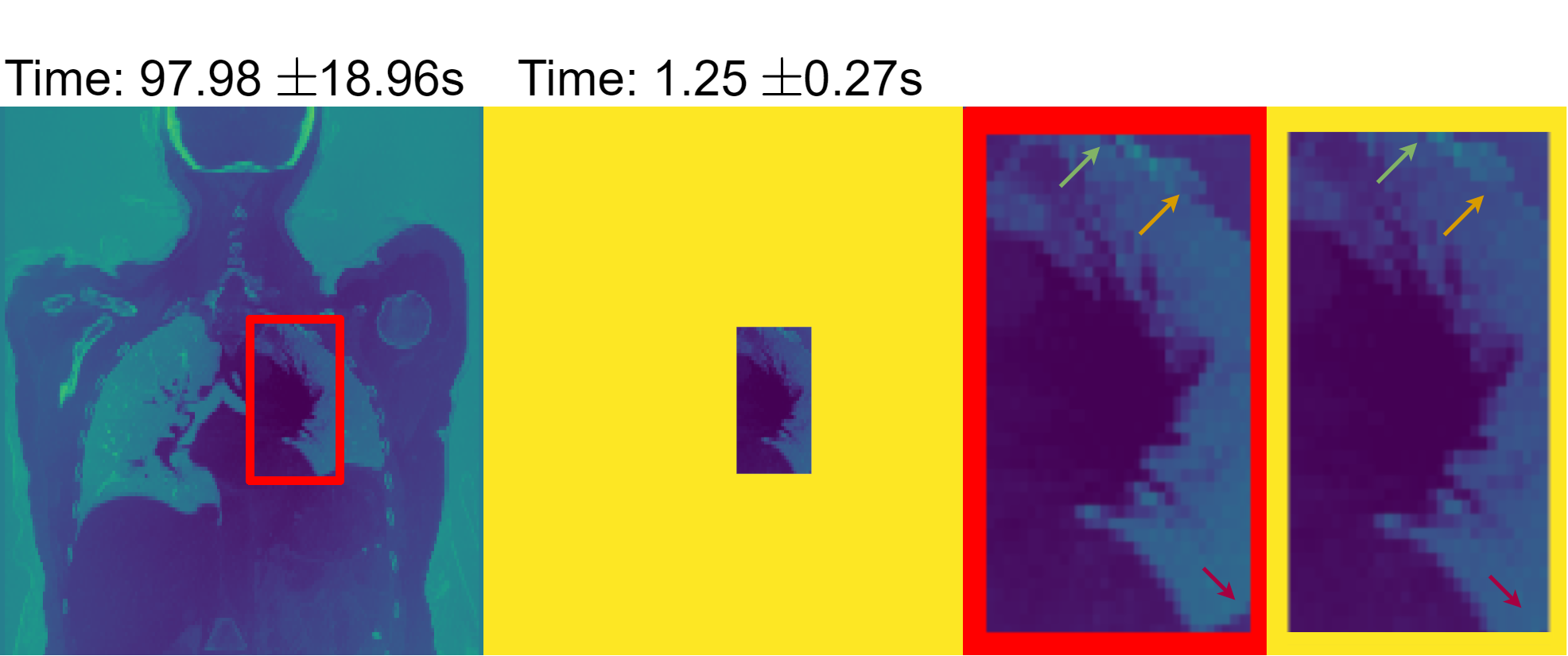}
    \caption{Qualitative and time comparison between the full-image GDT and the RoI GDT. Slight differences between the two are indicated by the arrows.}
    \label{fig:geodestic-comparison}
\end{figure}

\textbf{Training.} Since the input to P-Net consists only of the PET and CT volumes we train it end-to-end with the Dice loss. However, R-Net requires user input, which we simulate during training. We simulate each annotation as an ellipsoid of random center and axis sizes. Centers for foreground ellipsoid annotations are randomly sampled from the ground truth mask. Negative ellipsoid centers are sampled near the outer border of the ground truth mask since annotators typically click near the boundary when adding background clicks. To achieve this, we expand the ground-truth bounding box and sample negative centers from the expanded margin. Algorithm \ref{algo:user-input-sampling} describes this procedure. We calculate the geodesic distance transform (GDT) based on the sampled foreground and background ellipsoids and the CT image and then append it to the input for R-Net as well as the previous prediction. Additionally, we only use an RoI around the ellipsoid to calculate the GDT which leads to a 79x faster computation using the Raster-Scan algorithm~\cite{GeodesicRasterScan} as seen in Figure \ref{fig:geodestic-comparison}. The RoI GDT also preserves the details from the full-image GDT with a few small exceptions indicated by the arrows in Figure \ref{fig:geodestic-comparison}. 

\textbf{Inference.} During inference, we use the real user annotations as ground truth foreground and background voxels and calculate the RoI GDT based on them to append it to the PET/CT input. If there are no foreground/background scribbles we leave the GDT map empty.

\begin{algorithm}
\SetAlgoLined
\SetKwInput{Input}{Input}
\SetKwInput{Constants}{Constants}
\Input{{Image $\mathcal{I}\subset \mathbb{R}^{W \times H \times D}$}}
\Constants{{Axis scaling factor $\alpha$, Bounding box scaling factor $\beta$, Minimum axis size $m$}}
\KwResult{{Foreground or background ellipsoid $M$}}
\SetKwFunction{ellipsoid}{get\_random\_ellipsoid}
{
\If{Calculate foreground mask}{
    $\mathcal{P} \gets \text{\tt{foreground\_voxels}}(\mathcal{I})$\;
}
\Else{
    $\mathcal{B} \gets \beta \cdot \text{\tt{bbox}(\tt{foreground\_voxels}}(\mathcal{I}))$\;
    $\mathcal{P} \gets \operatorname{\tt{clip\_margin}}\left(\mathcal{B}\right)$\;
}
$\textbf{p} \gets \tt{random\_sample}(\mathcal{P}) $ \hspace{1cm} \tt{// center}\;
$\textbf{r} \sim \mathcal{U}\left(0, 1\right)^3$\;

$\textbf{a}[0] \gets \max\left(m, \textbf{r} \cdot \alpha \cdot W\right)$ \hspace{0.45cm} \tt{// x-axis}\;
$\textbf{a}[1] \gets \max\left(m, \textbf{r} \cdot \alpha \cdot H\right)$ \hspace{0.45cm} \tt{// y-axis}\;
$\textbf{a}[2] \gets \max\left(m,\textbf{r} \cdot \alpha \cdot D\right)$ \hspace{0.45cm} \tt{// z-axis}\;
$M \gets \operatorname{\tt{calc\_ellipsoid}}\left(\textbf{a}, \textbf{p}\right)$\;
\Return $M$\;
}

\caption{\tt{sample\_user\_input}}\label{algo:user-input-sampling}
\end{algorithm}

\section{Setup and Results}

\subsection{Training and Implementation Details}
The dataset used for training and evaluation originates from the \emph{Automated Lesion Segmentation in Whole-Body PET/CT Challenge} (AutoPET)~\cite{AutoPET}. We select only patients with lung cancer lesions, which resulted in 122 studies for training and validation and 30 studies for testing. For the training of P-Net and R-Net, we apply random rotations with a probability of 0.5 for each axis and a range of [-10, 10] degrees, random affine transformation with a translation range of 10\% and probability of 30\%, and random Gaussian Noise with a probability of 30\%. For optimization, Adam~\cite{kingma2014adam} is used with a batch size of 1. The models are trained for 36k iterations.
We utilize the user simulation scheme described in Algorithm \ref{algo:user-input-sampling} to generate foreground and background annotations. We apply our RoI geodesic distance transform to these annotations and concatenate them to the input during the training of R-Net. The simulated user foreground annotations range between one and three samples, whereas the background annotations can be either zero or one.


\begin{figure*}[t]
    \centering
    \includegraphics[width=\textwidth]{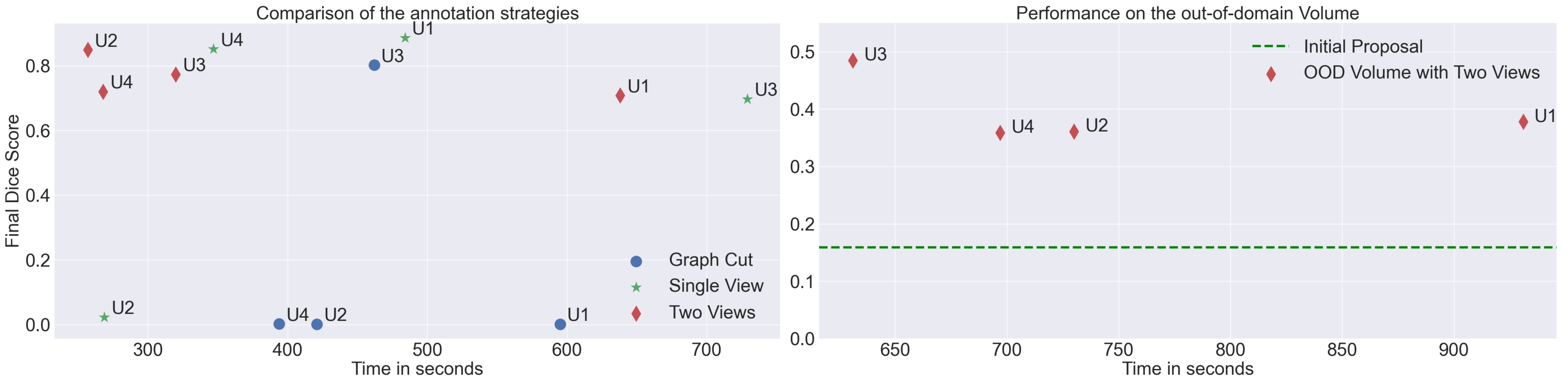}
    \caption{Results from our user study. Each point represents the time and dice score achieved by the corresponding user (U*).}
    \label{fig:scatter_plots}
\end{figure*}

\subsection{User Study Setup}\label{subsec:study-setup}

The model is evaluated 
through a user study involving four trained medical experts.
The goal is to validate the following three hypotheses:
(1) showing both modalities simultaneously in the user interface accelerates the annotation; (2)
the annotation using our proposed model with two views outperforms its single view counterpart and GraphCut~\cite{GraphCutCalculation}; (3)
our model improves the initial proposal on unseen data.

To investigate all hypotheses, the study was performed as an A/B-test. The participants annotated four volumes, each in one of three different settings, with the order of settings being cycled for each participant. The first setting utilized our model and a two-view user interface where PET in CT data is displayed side-by-side during annotation. The second setting used GraphCut~\cite{GraphCutCalculation} and our proposed two-view user interface. The last setting utilized our model and a user interface that shows one modality at a time, with the users being able to switch the modality using a button, as shown in Figure \ref{fig:interface} on the right.

The volumes were chosen from the test split and contain lesions of approximately the same size, with three being from the AutoPET~\cite{AutoPET} dataset and one from the Lung-PET-CT-Dx dataset~\cite{LungPETCT}.
The last volume was always annotated with the first setting as it is from an unseen dataset. 


\subsection{Evaluation}\label{subsec:evaluation}

Figure \ref{fig:scatter_plots} visualizes the results from our user study. All users utilizing our method (Two Views) have achieved an acceptable Dice score on their volume. In contrast, the majority of users using GraphCut \cite{GraphCutAlgorithm} could not produce adequate annotations leading to close to zero Dice scores. Users using the single view interface achieved similar Dice scores to the two parallel views, with the exception of User 2. However, the annotation time with the single view is considerably slower. Hence, our two-views interface is both faster and produces better results than the other methods, proving our hypotheses (1) and (2). Figure \ref{fig:scatter_plots} also shows the performance of each user on an unseen volume, using our two-view annotation interface. The results demonstrate that each user significantly improved the initial segmentation from P-Net using our interface, which confirms our last hypothesis (3).

In a post-study questionnaire, the participants considered viewing both modalities as very helpful and the speed of the proposal as very acceptable. On average, the annotation process using our model took approximately one minute less per volume than when using GraphCut, expediting the process by approximately 12.5\%. 
The participants described themselves as having little to no experience with 3D Slicer and the annotation of PET/CT images. Nevertheless, they were able to produce high-quality segmentations on AutoPET \cite{AutoPET} and improve the automatic segmentation on unseen volumes.

\section{Discussion and Conclusion}
In this work, we adapt the concept of interactive segmentation to multi-modal images and propose a novel framework for PET/CT volume annotation. 
Subsequently, we extend DeepIGeoS~\cite{wang2018deepigeos} for multi-modal imaging data. To simulate user interactions, we utilize random ellipsoids and encode interactions using an efficient RoI geodesic distance transform. 

We investigated our method for the annotation of non-small-cell lung carcinoma.
We have displayed the effects of different views of multi-modal PET/CT images on the annotation process with the results indicating that the joined view of both modalities typically leads to better and faster annotation. 
Our user study shows that using our proposed annotation framework experts can save on average several minutes in annotation time per volume compared to other setups. 

We publicly release our code to not only foster future research in the field of multi-modal interactive segmentation but also enable an easier generation of medical datasets. 

Future work will focus on improving the user interaction with the system.
Besides smaller user interface adjustments suggested by the user study participants, the system could guide the user annotations by proposing slices where the model is the most uncertain in its prediction, as proposed by Wang et~al.~\cite{wang2020uncertainty}.
This can further help in reducing the total time required by the users for their annotation tasks.



\section{Compliance with Ethical Standards}
This research study was conducted retrospectively using human subject data made available in open access. Ethical approval was *not* required as confirmed by the license attached with the open access data.
\section{Acknowledgments}
\label{sec:acknowledgments}
The present contribution is supported by the Helmholtz Association under the joint research school “HIDSS4Health – Helmholtz Information and Data Science School for Health.”

\bibliographystyle{IEEEbib}
\bibliography{strings,refs}

\begin{thebibliography}{10}

\bibitem{russakovsky2015imagenet}
Olga Russakovsky et~al.,
\newblock ``Imagenet large scale visual recognition challenge,''
\newblock {\em International journal of computer vision}, vol. 115, no. 3, pp.
  211--252, 2015.

\bibitem{patel2020upsurge}
Priyanka Patel and Amit Thakkar,
\newblock ``The upsurge of deep learning for computer vision applications,''
\newblock {\em International Journal of Electrical and Computer Engineering},
  vol. 10, no. 1, pp. 538, 2020.

\bibitem{dosovitskiy2020image}
Alexey Dosovitskiy et~al.,
\newblock ``An image is worth 16x16 words: Transformers for image recognition
  at scale,''
\newblock {\em arXiv preprint arXiv:2010.11929}, 2020.

\bibitem{ronneberger2015u}
Olaf Ronneberger et~al.,
\newblock ``U-net: Convolutional networks for biomedical image segmentation,''
\newblock in {\em International Conference on Medical image computing and
  computer-assisted intervention}. Springer, 2015, pp. 234--241.

\bibitem{rother2021assessing}
Anne Rother et~al.,
\newblock ``Assessing the difficulty of annotating medical data in crowdworking
  with help of experiments,''
\newblock {\em PloS one}, vol. 16, no. 7, pp. e0254764, 2021.

\bibitem{wang2018deepigeos}
Guotai Wang et~al.,
\newblock ``Deepigeos: a deep interactive geodesic framework for medical image
  segmentation,''
\newblock {\em IEEE transactions on pattern analysis and machine intelligence},
  vol. 41, no. 7, pp. 1559--1572, 2018.

\bibitem{luo2021mideepseg}
Xiangde Luo et~al.,
\newblock ``Mideepseg: Minimally interactive segmentation of unseen objects
  from medical images using deep learning,''
\newblock {\em Medical Image Analysis}, vol. 72, pp. 102102, 2021.

\bibitem{wang2020uncertainty}
Guotai Wang et~al.,
\newblock ``Uncertainty-guided efficient interactive refinement of fetal brain
  segmentation from stacks of mri slices,''
\newblock in {\em International Conference on Medical Image Computing and
  Computer-Assisted Intervention}. Springer, 2020, pp. 279--288.

\bibitem{roth2021going}
Holger~R Roth et~al.,
\newblock ``Going to extremes: weakly supervised medical image segmentation,''
\newblock {\em Machine Learning and Knowledge Extraction}, vol. 3, no. 2, pp.
  507--524, 2021.

\bibitem{sommer2011ilastik}
Christoph Sommer et~al.,
\newblock ``Ilastik: Interactive learning and segmentation toolkit,''
\newblock in {\em 2011 IEEE international symposium on biomedical imaging: From
  nano to macro}. IEEE, 2011, pp. 230--233.

\bibitem{cho2021deepscribble}
Sungduk Cho et~al.,
\newblock ``Deepscribble: interactive pathology image segmentation using deep
  neural networks with scribbles,''
\newblock in {\em 2021 IEEE 18th International Symposium on Biomedical Imaging
  (ISBI)}. IEEE, 2021, pp. 761--765.

\bibitem{li2021wdtiseg}
Xiaokang Li et~al.,
\newblock ``Wdtiseg: One-stage interactive segmentation for breast ultrasound
  image using weighted distance transform and shape-aware compound loss,''
\newblock {\em Applied Sciences}, vol. 11, no. 14, pp. 6279, 2021.

\bibitem{AutoPET}
Sergios Gatidis, Tobias Hepp, Marcel Fr{\"u}h, Christian La~Foug{\`e}re,
  Konstantin Nikolaou, Christina Pfannenberg, Bernhard Sch{\"o}lkopf, Thomas
  K{\"u}stner, Clemens Cyran, and Daniel Rubin,
\newblock ``A whole-body fdg-pet/ct dataset with manually annotated tumor
  lesions,''
\newblock {\em Scientific Data}, vol. 9, no. 1, pp. 1--7, 2022.

\bibitem{krause2013fdg}
Berud~J Krause et~al.,
\newblock ``Fdg pet and pet/ct,''
\newblock {\em Molecular Imaging in Oncology}, pp. 351--369, 2013.

\bibitem{townsend2004pet}
David~W Townsend et~al.,
\newblock ``Pet/ct today and tomorrow,''
\newblock {\em Journal of Nuclear Medicine}, vol. 45, no. 1 suppl, pp. 4S--14S,
  2004.

\bibitem{chen2004application}
Yen-Kung Chen et~al.,
\newblock ``Application of pet and pet/ct imaging for cancer screening,''
\newblock {\em Anticancer research}, vol. 24, no. 6, pp. 4103--4108, 2004.

\bibitem{LungPETCT}
Ping Li et~al.,
\newblock ``A large-scale ct and pet/ct dataset for lung cancer diagnosis,''
  2020.

\bibitem{Slicer}
Andriy Fedorov et~al.,
\newblock ``3d slicer as an image computing platform for the quantitative
  imaging network,''
\newblock {\em Magnetic Resonance Imaging}, vol. 30, no. 9, pp. 1323--1341,
  2012,
\newblock Quantitative Imaging in Cancer.

\bibitem{MONAILabel}
Andres Diaz-Pinto et~al.,
\newblock ``Monai label: A framework for ai-assisted interactive labeling of 3d
  medical images,'' 2022.

\bibitem{GeodesicRasterScan}
Antonio Criminisi et~al.,
\newblock ``Geos: Geodesic image segmentation,''
\newblock in {\em Computer Vision -- ECCV 2008}, Berlin, Heidelberg, 2008, pp.
  99--112, Springer Berlin Heidelberg.

\bibitem{kingma2014adam}
Diederik~P Kingma and Jimmy Ba,
\newblock ``Adam: A method for stochastic optimization,''
\newblock {\em arXiv preprint arXiv:1412.6980}, 2014.

\bibitem{GraphCutCalculation}
Guotai Wang et~al.,
\newblock ``Interactive medical image segmentation using deep learning with
  image-specific fine tuning,''
\newblock {\em {IEEE} Transactions on Medical Imaging}, vol. 37, no. 7, pp.
  1562--1573, jul 2018.

\bibitem{GraphCutAlgorithm}
Y.~Boykov and V.~Kolmogorov,
\newblock ``An experimental comparison of min-cut/max- flow algorithms for
  energy minimization in vision,''
\newblock {\em {IEEE} Transactions on Pattern Analysis and Machine
  Intelligence}, vol. 26, no. 9, pp. 1124--1137, sep 2004.

\end{thebibliography}

\end{document}